\pdfoutput=1

\documentclass[11pt]{article}

\usepackage[]{acl}

\usepackage{times}
\usepackage{latexsym}

\usepackage[T1]{fontenc}

\usepackage[utf8]{inputenc}

\usepackage{microtype}

\usepackage[pdftex]{graphicx}
\usepackage{subcaption}
\usepackage{booktabs}
\usepackage{multirow}

\title{Federated Named Entity Recognition}

\author{Joel Mathew, Dimitris Stripelis \and Jos\'{e} Luis Ambite \\
        USC Information Sciences Institute}

\begin{document}
\maketitle
\begin{abstract}

We present an analysis of the performance of Federated Learning in a paradigmatic natural-language processing task: Named-Entity Recognition (NER). For our evaluation, we use the language-independent CoNLL-2003 dataset as our benchmark dataset and a Bi-LSTM-CRF model as our benchmark NER model. We show that federated training reaches almost the same performance as the centralized model, though with some performance degradation as the learning environments become more heterogeneous. We also show the convergence rate of federated models for NER. Finally, we discuss existing challenges of Federated Learning for NLP applications that can foster future research directions.


\end{abstract}

\section{Introduction}
To protect personal and proprietary data from illegal access and malicious use, new data regulatory frameworks (e.g., GDPR, CCPA, PIPL) have been recently enacted that govern data access, processing and analysis. When also taking into account the increasing amount of data being generated across various domains, new tools are required to perform meaningful large scale data analysis while at the same complying to the newly introduced data privacy and security legislations.

One potential solution towards secure and private data analysis is Federated Learning~\cite{mcmahan2017communication,kairouz2021advances,yang2019federated}. Federated Learning has emerged as a standard computational paradigm for distributed machine learning that allows geographically distributed data sources to efficiently train machine learning models while their raw data remain at their original location. Compared to traditional machine learning algorithms that require all training data to be aggregated at a centralized location, Federated Learning relaxes the need of raw data migration and instead pushes the training of the machine learning model down to each source. During federated training only the locally trained model parameters are shared with the outside world and a centralized entity is responsible to aggregate these parameters to compute the federation model. 

Due to its inherent data security and privacy benefits, Federated Learning has seen wide adoption across many different disciplines and applications ranging from smart transportation~\cite{liu2020privacy}, finance~\cite{liu2020fedcoin} and healthcare~\cite{sheller2020federated,rieke2020future}. In this work, we study Federated Learning in a more generalized setting within the context of Natural Language Processing (NLP) with a focus on the Named Entity Recognition (NER) task. 

NER is an important tool for textual analysis that has been applied across various domains, including information extraction, question-answering, and semantic annotation~\cite{marrero2013named}. However, some domains require the textual data to remain private; for example, intelligence applications or analyses of email data across different organizations. These private-text domains motivate our research into federated learning approaches for NLP problems, such as Named Entity Recognition.




\section{Related Work}
Federated Learning was originally introduced in the seminal work of McMahan et al. (\citeyear{mcmahan2017communication}) for user data in mobile phones. In their work, the authors proposed the first synchronous federated learning algorithm, termed Federated Average (FedAvg), where a set of clients is randomly selected at every federation round to train the global model on their local dataset under the coordination of a central server who is responsible to aggregate the local models into the global model and distribute the respective local training workload to every client. Following this learning approach, recently many other federated training approaches~\cite{kairouz2021advances,li2020federated,yang2019federated} have been proposed both for mobile devices (cross-device) and large-scale organizations (cross-silo). 

Other approaches have provided convergence guarantees for the original FedAvg algorithm over non-IID data~\cite{li2019convergence} while others have decoupled the federated optimization problem into global and local optimization~\cite{wang2021field,reddi2020adaptive} with linear~\cite{mitra2021linear,karimireddy2020scaffold} and sub-linear convergence guarantees~\cite{li2020federated,wang2020tackling} in the presence of different client learning constraints, such as system (different computational capabilities) and statistical heterogeneities (different local dataset distributions).

Even though many federated training approaches have been proposed most of them focus in the context of computer vision and only a handful of them provide a federated solution tailored for NLP applications~\cite{liu2021federated,lin2021fednlp}. For instance, federated learning has been applied to tackle the problem of next word keyboard prediction~\cite{hard2018federated,stremmel2021pretraining,yang2018applied}, speech recognition~\cite{leroy2019federated} and health text mining~\cite{liu2019two}. In the context of sequence tagging and named entity recognition recent works have investigated the feasibility of a federated learning solution on medical named entity extraction~\cite{ge2020fedner} and extraction of personally identifiable information elements from text documents~\cite{hathurusinghe2021privacy}.

For the NER task that we investigate in this work, one of the first successful uses of neural networks was the Bi-LSTM~\cite{Hochreiter1997LongSM} and CRF~\cite{Lafferty2001ConditionalRF} models, with both models combined into a single state-of-the-art deep learning model architecture~\cite{lample-etal-2016-neural} that exhibited superior performance. Other recent works have proposed large transformer-based models such as BERT~\cite{devlin-etal-2019-bert} and XLM~\cite{Lample2019CrosslingualLM}. However, for understanding the impact of federated training in the learning performance of the NER task, in this work we employ the Bi-LSTM-CRF model, similar to~\cite{mathew2019biomedical}.

\section{NER Model}
\begin{figure}[htbp]
  \centering
  \includegraphics[width=\columnwidth]{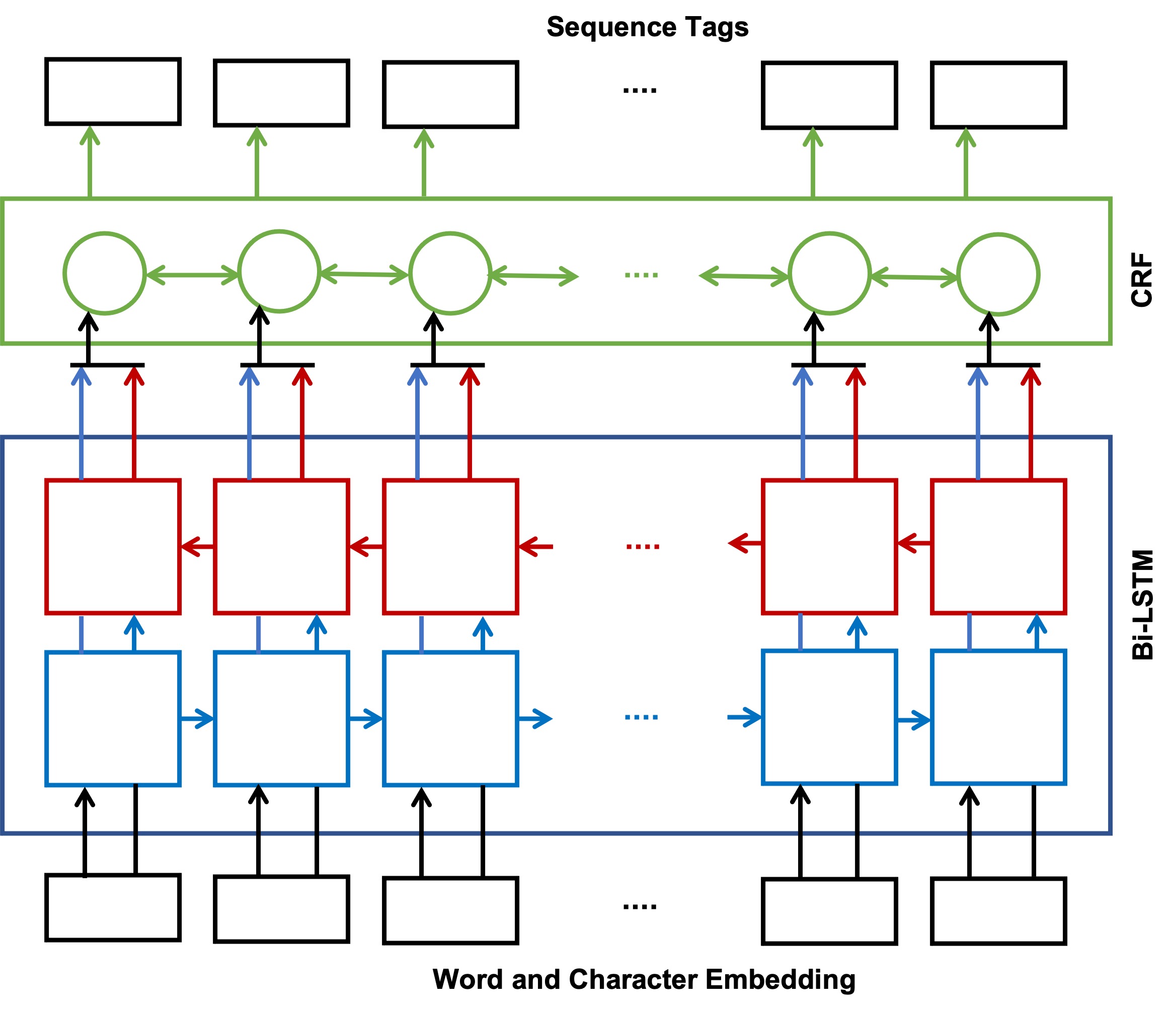}
   \caption{NER Model Architecture.}
   \label{fig:model}
\end{figure}

We use a Bi-LSTM layer which is fed a concatenation of 300-dimension GloVe ~\cite{pennington-etal-2014-glove} word embedding and a character embedding that is trained on the data as input. We use dropout during training, set at 0.5. The output of the Bi-LSTM model is then fed into a CRF layer in order to capture neighboring tagging decisions. The CRF layer produces scores for all possible sequences of tags over which we apply the Softmax function to produce the output tag sequence. Figure~\ref{fig:model} shows the architecture of the deep learning model used throughout training.%
\footnote{\url{https://github.com/guillaumegenthial/tf\_ner}}
 The total number of trainable parameters are 322,690.

\section{Experiments}
For all learning environments (centralized and federated) the random seed was set to 1990. We use Vanilla SGD as the solver of the centralized model and the local model in the federation with a learning rate of 0.01 and batch size of 20.

\textbf{Federated Training.}
Our federated environments follow a centralized learning topology~\cite{bellavista2021decentralised,yang2019federated,kairouz2021advances} where a single aggregator (server) is responsible to orchestrate the execution of the participating clients. In our experiments, we consider full client participation at every round. We test the performance of the federated model on federation environments consisting of 8, 16 and 32 clients. Each client trained on its local dataset for 4 local epochs in-between federation rounds and each federated experiment was run for a total number of 200 rounds. When merging the local models at the server, we used FedAvg as the merging function. During training clients shared the kernel and bias matrices of the LSTM and dense layers and the transition matrix of the CRF layer. All federated environments were run using the Metis framework~\cite{stripelis2021semi}.

\textbf{Federated Data Distributions.}
The CoNLL-2003~\cite{sang2003introduction} is a language-independent newswire dataset developed for the named entity recognition task. The dataset consists of 20,744 sentences (14041 training, 3250 validation, 3453 test) and contains entities referring to locations (LOC), organizations (ORG), people (PER), and miscellaneous (MISC).~\footnote{Total number of entities in the training and validation sets combined, LOC: 8977, ORG: 7662, PER: 8442, MISC: 4360. For test, LOC: 1668, ORG: 1661, PER: 1617, MISC: 702} We measure that classification performance for all entities. The original tagging scheme is BIO (beginning-intermediate-other). When extracting the named entities two subtasks need to be solved, finding the exact boundaries of an entity, and the entity type. The metrics used to evaluate correct tag (entity) predictions are Precision, Recall and F1. All models were evaluated on the same original test dataset. 

\begin{figure}[htbp]
  \captionsetup[subfigure]{justification=centering}
  \centering
  \subfloat[8-Clients\\Uniform]{
    \centering\includegraphics[width=0.5\linewidth]{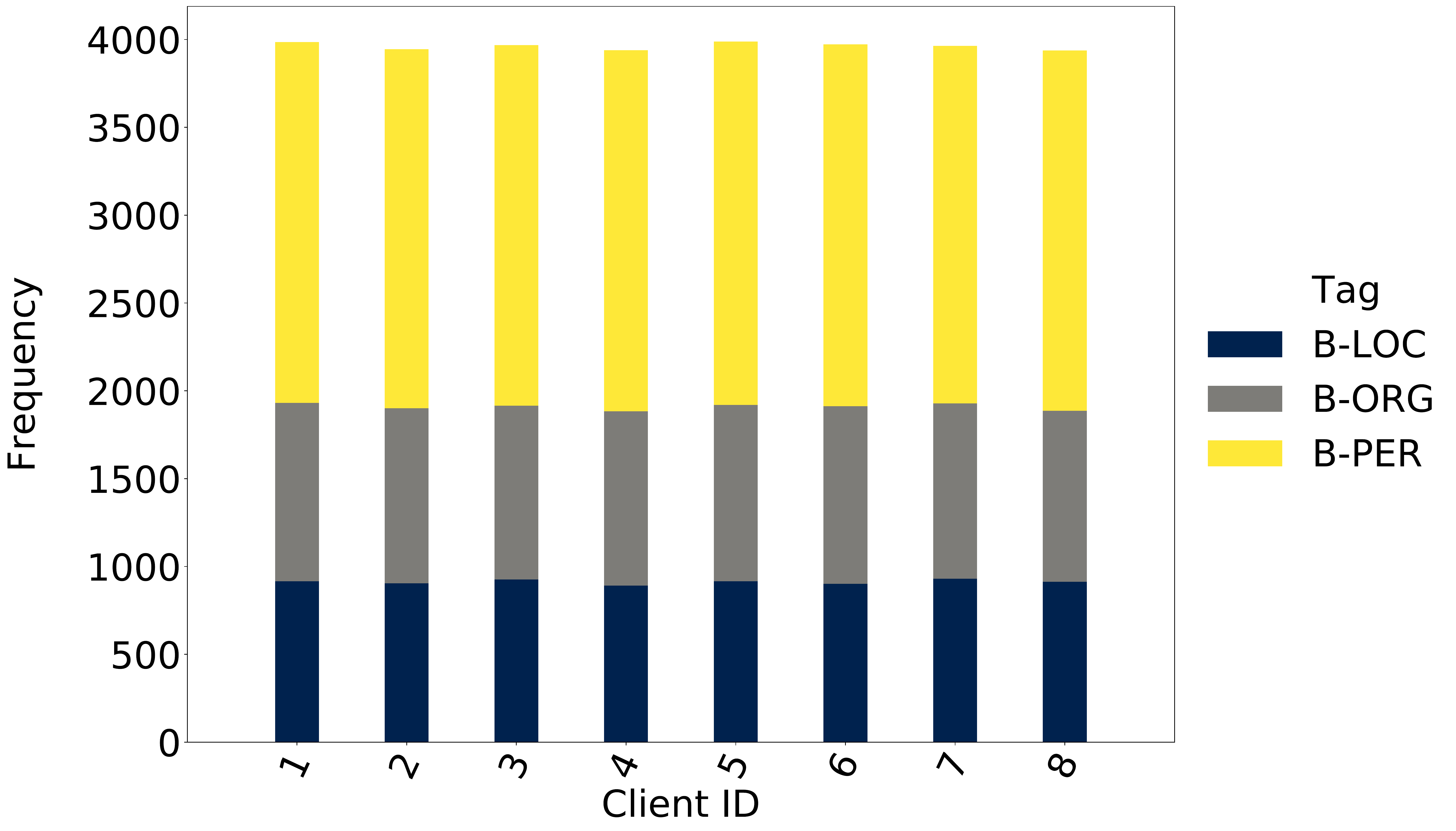}
    \label{subfig:}
  }
  \subfloat[16-Clients\\Uniform]{
    \centering\includegraphics[width=0.5\linewidth]{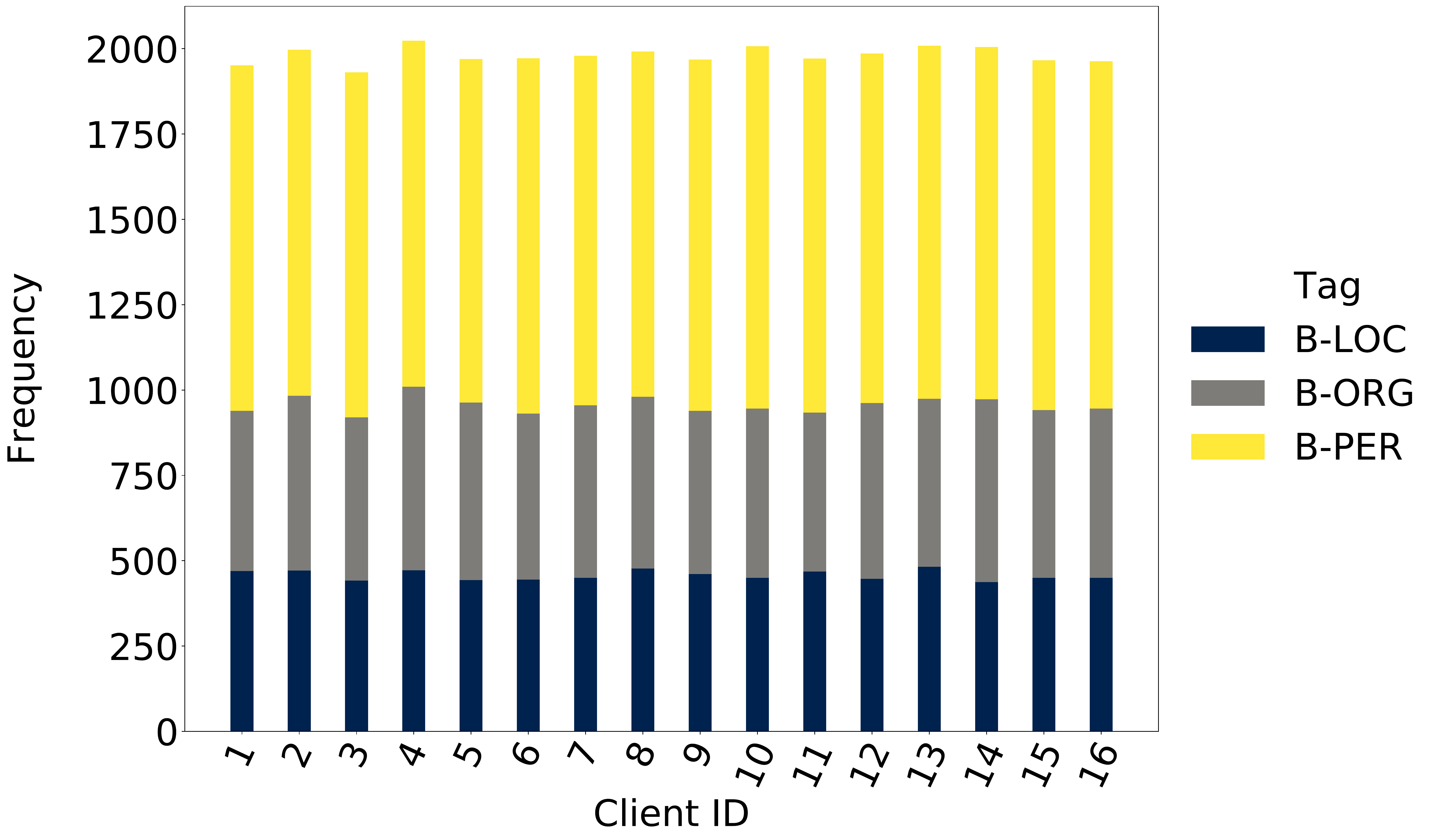}
    \label{subfig:}
  }
  
  \subfloat[32-Clients\\Uniform]{
    \centering\includegraphics[width=0.5\linewidth]{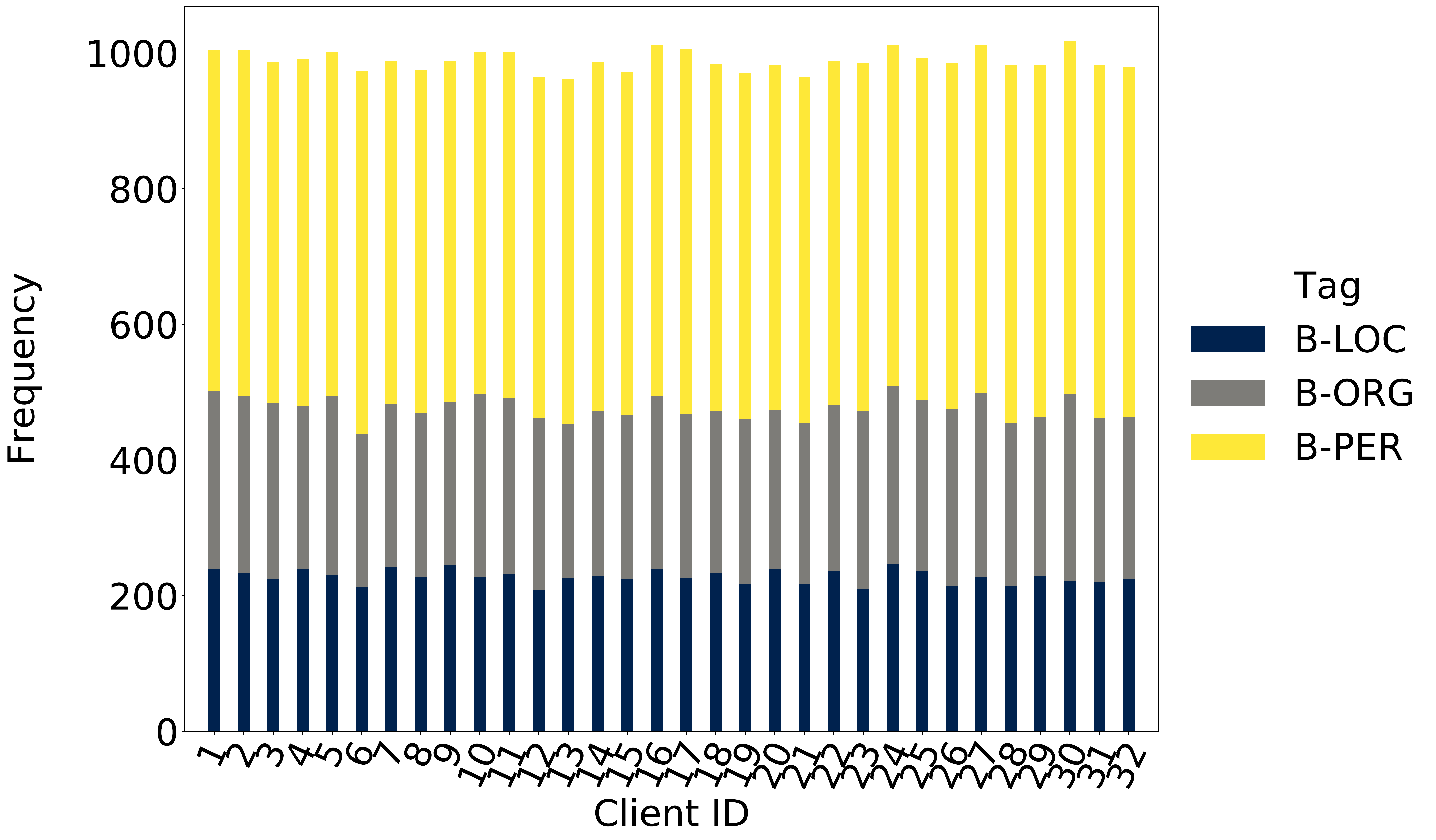}
    \label{subfig:}
  }
    \subfloat[32-Clients\\Skewed]{
    \centering\includegraphics[width=0.5\linewidth]{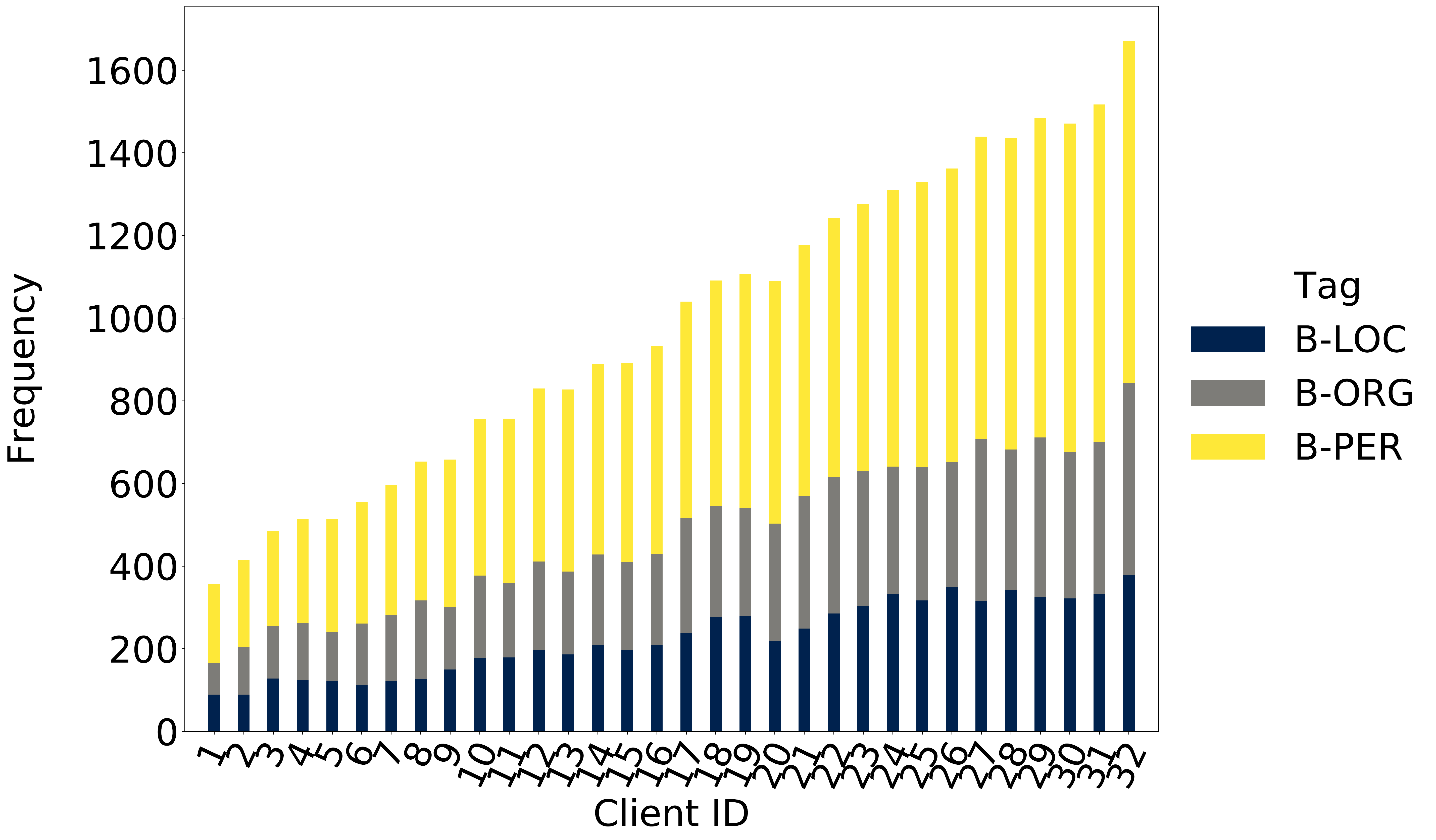}
    \label{subfig:}
  }
  \captionsetup{}
   \caption{CoNLL-2003 dataset: B-LOC, B-ORG and B-PER entity (tag) distribution for each client within each federation environment.}
  \label{fig:CoNLL_2003_tags_distribution}
\end{figure}

\begin{figure}[htbp]
  \captionsetup[subfigure]{justification=centering}
  \centering
  \subfloat[8-Clients\\Uniform]{
    \centering\includegraphics[width=0.5\linewidth]{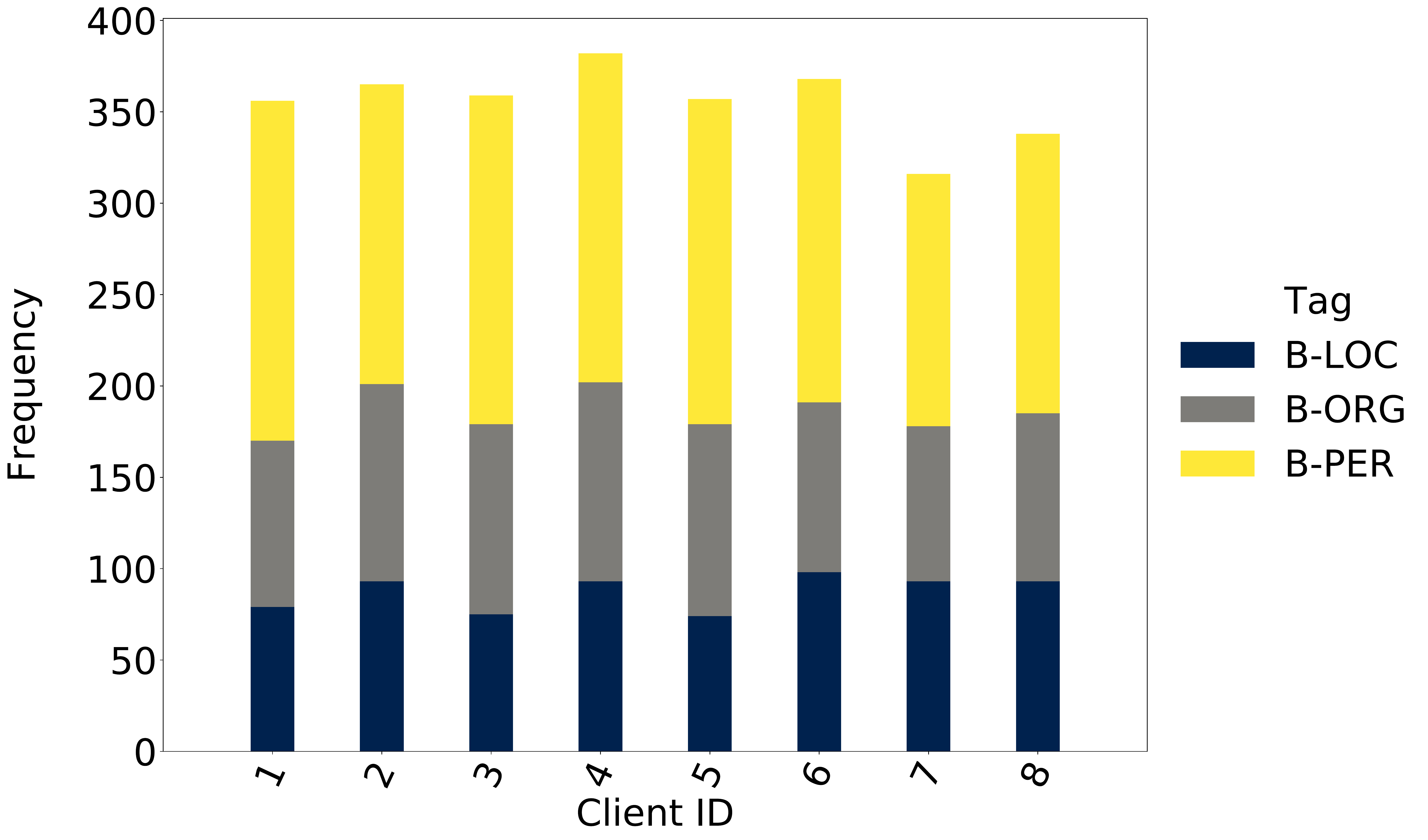}
    \label{subfig:}
  }
  \subfloat[16-Clients\\Uniform]{
    \centering\includegraphics[width=0.5\linewidth]{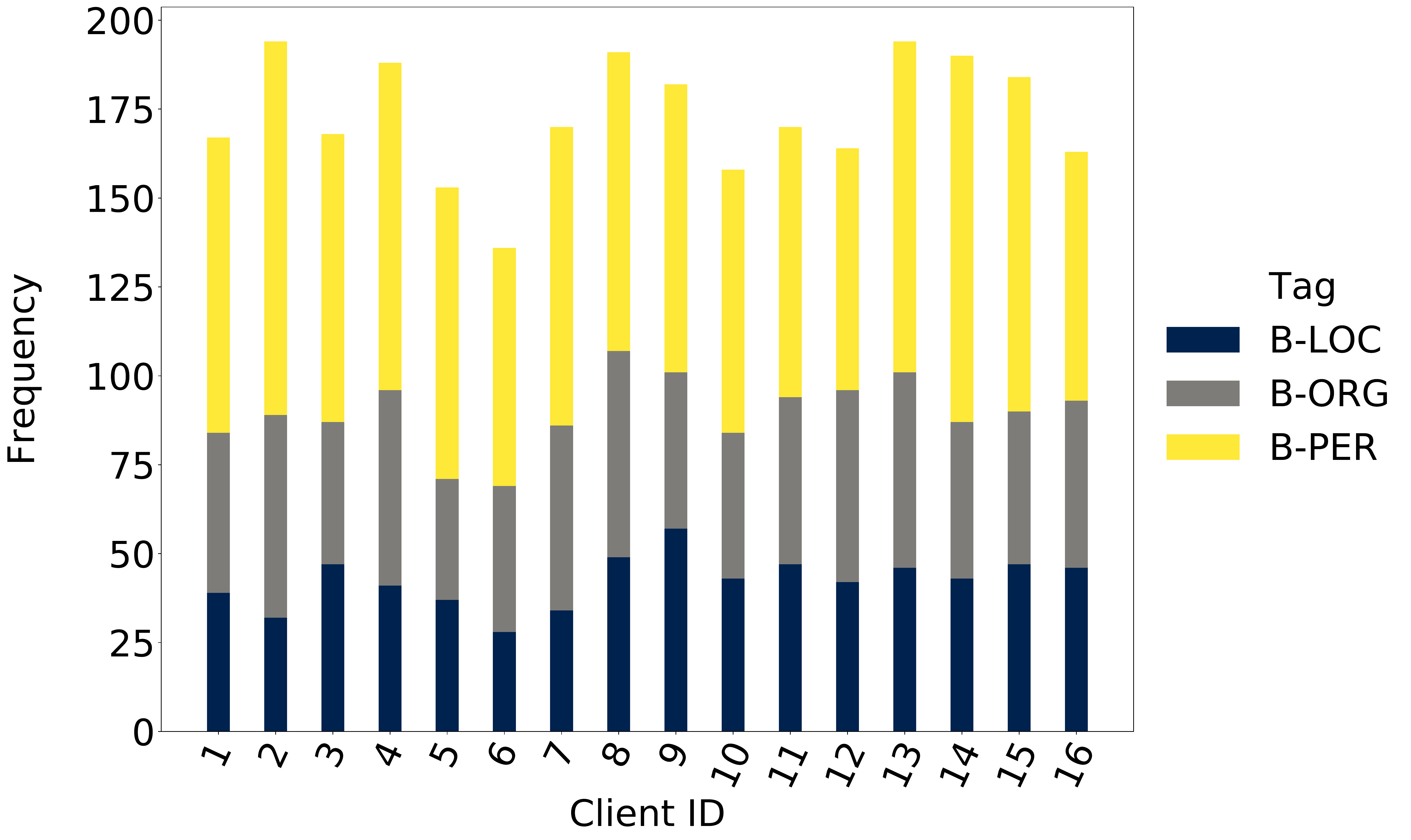}
    \label{subfig:}
  }
  
  \subfloat[32-Clients\\Uniform]{
    \centering\includegraphics[width=0.5\linewidth]{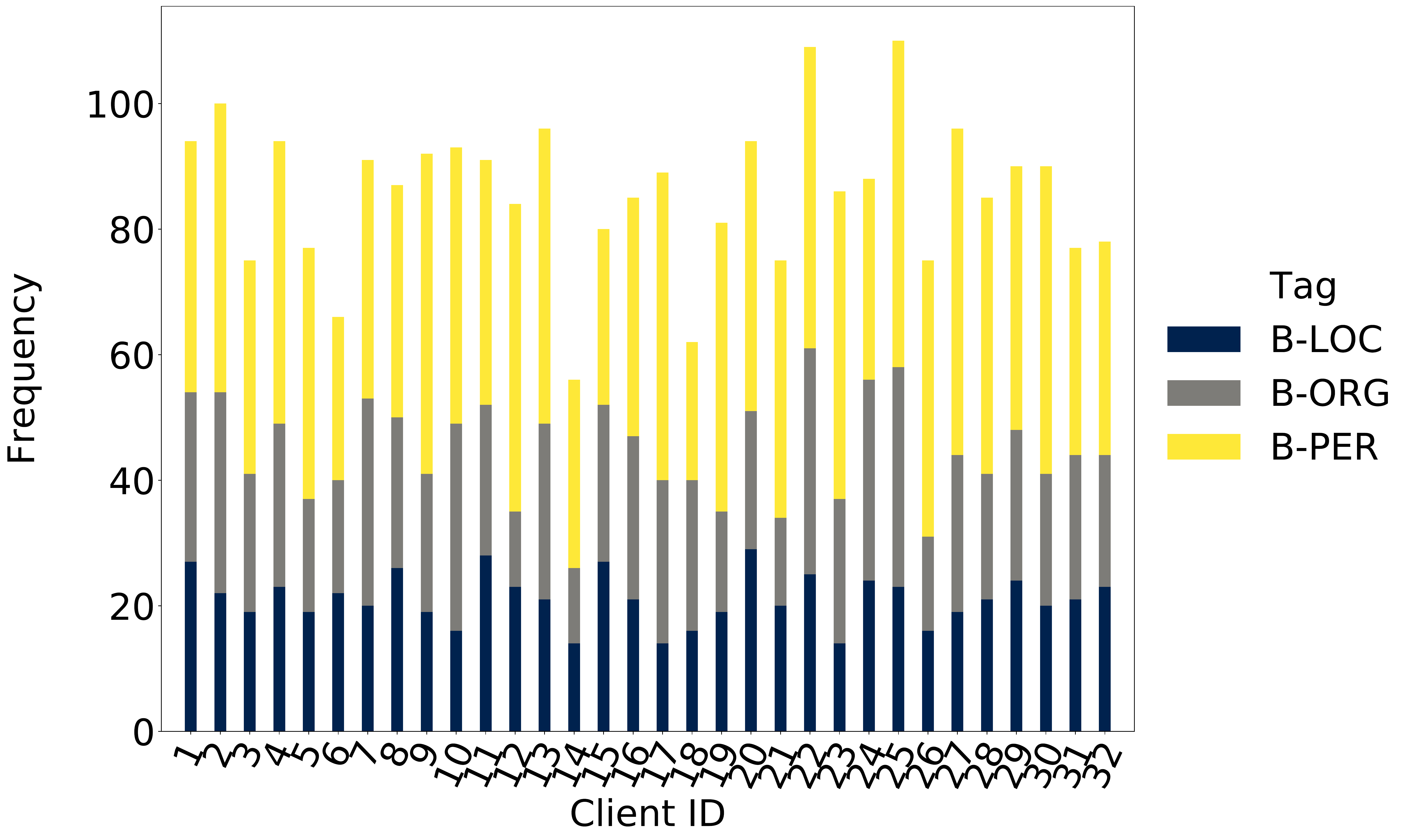}
    \label{subfig:}
  }
  \subfloat[32-Clients\\Skewed]{
    \centering\includegraphics[width=0.5\linewidth]{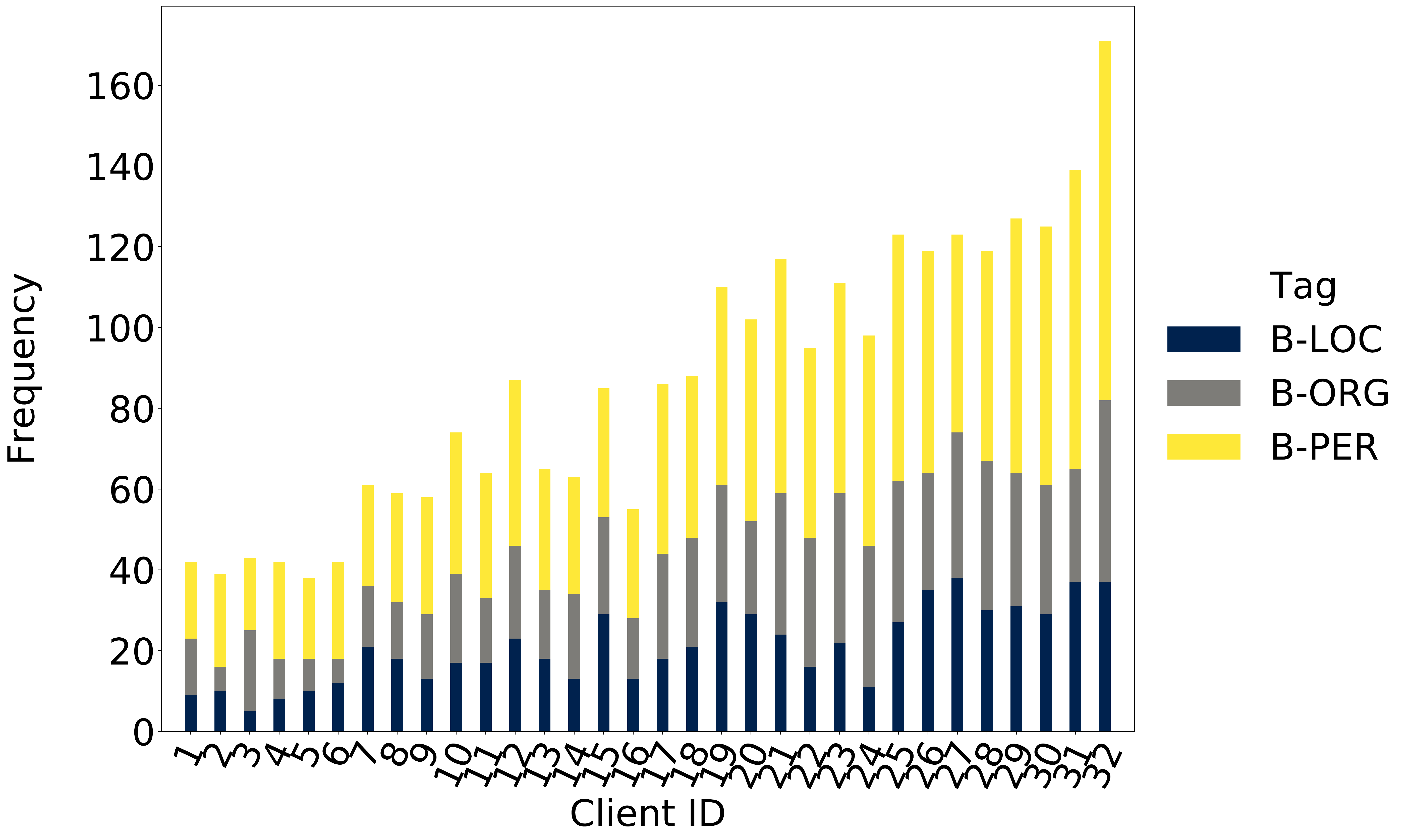}
    \label{subfig:non_uniform}
  }  
  \captionsetup{}
   \caption{CoNLL-2003 dataset: Number of \textbf{unique} B-LOC, B-ORG and B-PER entity (tag) distribution for each client within each federation environment.}
  \label{fig:CoNLL_2003_unique_tags_distribution}
\end{figure}

To generate the federated environments that we investigate in this work, we split the dataset into equal (Uniform) and unequal (Skewed) sized partitions. For the Uniform enviroments, we combine the training and validation datasets of the original dataset and split them into approximately equally sized partitions for 8, 16 and 32-clients, such that each partition (client) has almost the same proportion of different tag types. The proportion of tags in each split is approximate, as it can also be seen in Figure~\ref{fig:CoNLL_2003_tags_distribution}, since any sentence can have any number of different tags. For the 32-clients and Skewed environment, we randomly partitioned the combined training and validation datasets into 32 splits over an increasing amount of data points (200 to 872 sentences). This increased the data heterogeneity across all partitions with certain partitions containing more unique entity mentions (compared to the Uniform environments); see also Figure~\ref{subfig:non_uniform}.

\begin{figure}[htbp]
  \captionsetup[subfigure]{justification=centering}
  \centering
  \subfloat[8-Clients\\Uniform]{
    \centering\includegraphics[width=0.5\linewidth]{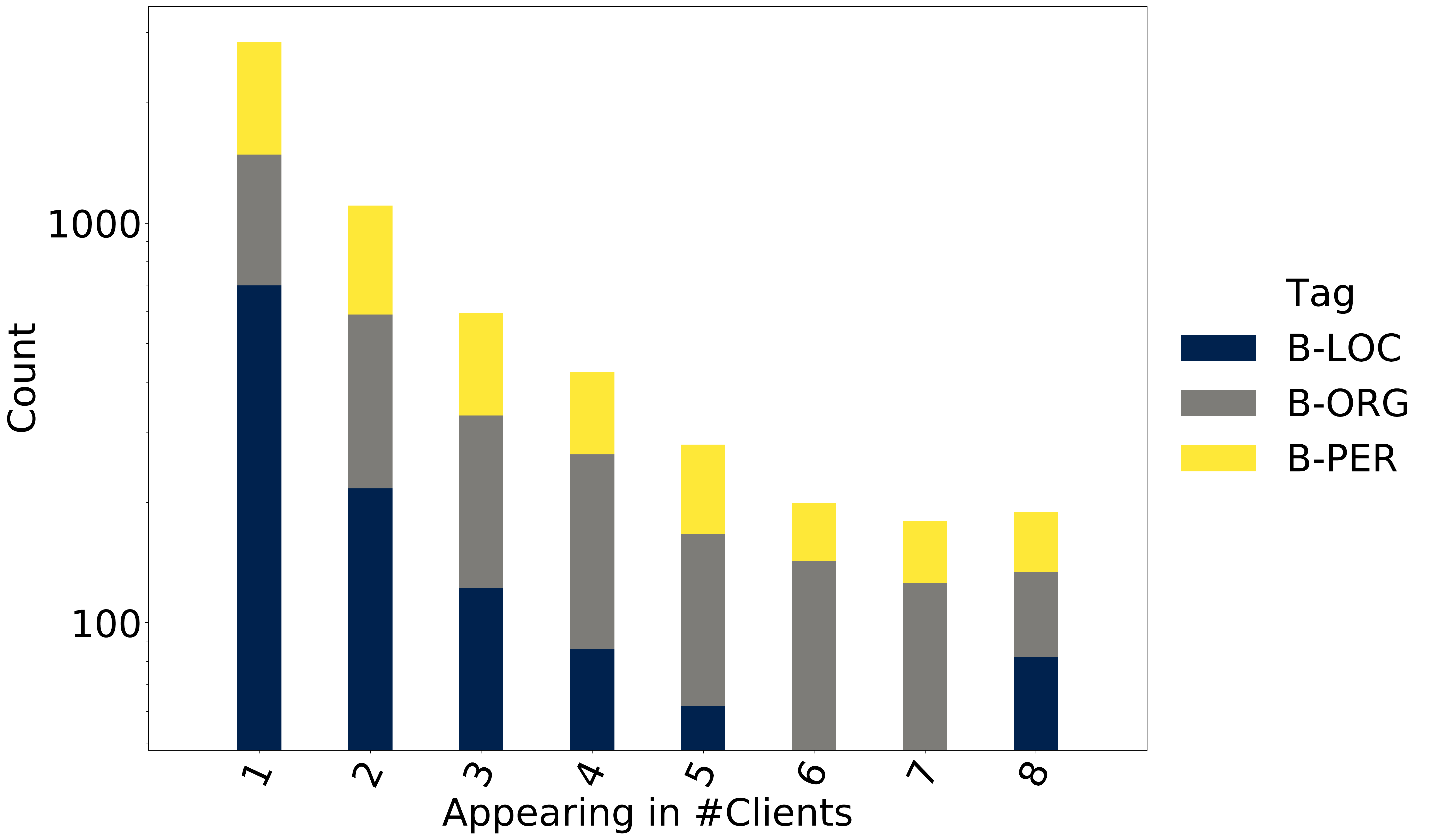}
    \label{subfig:}
  }
  \subfloat[16-Clients\\Uniform]{
    \centering\includegraphics[width=0.5\linewidth]{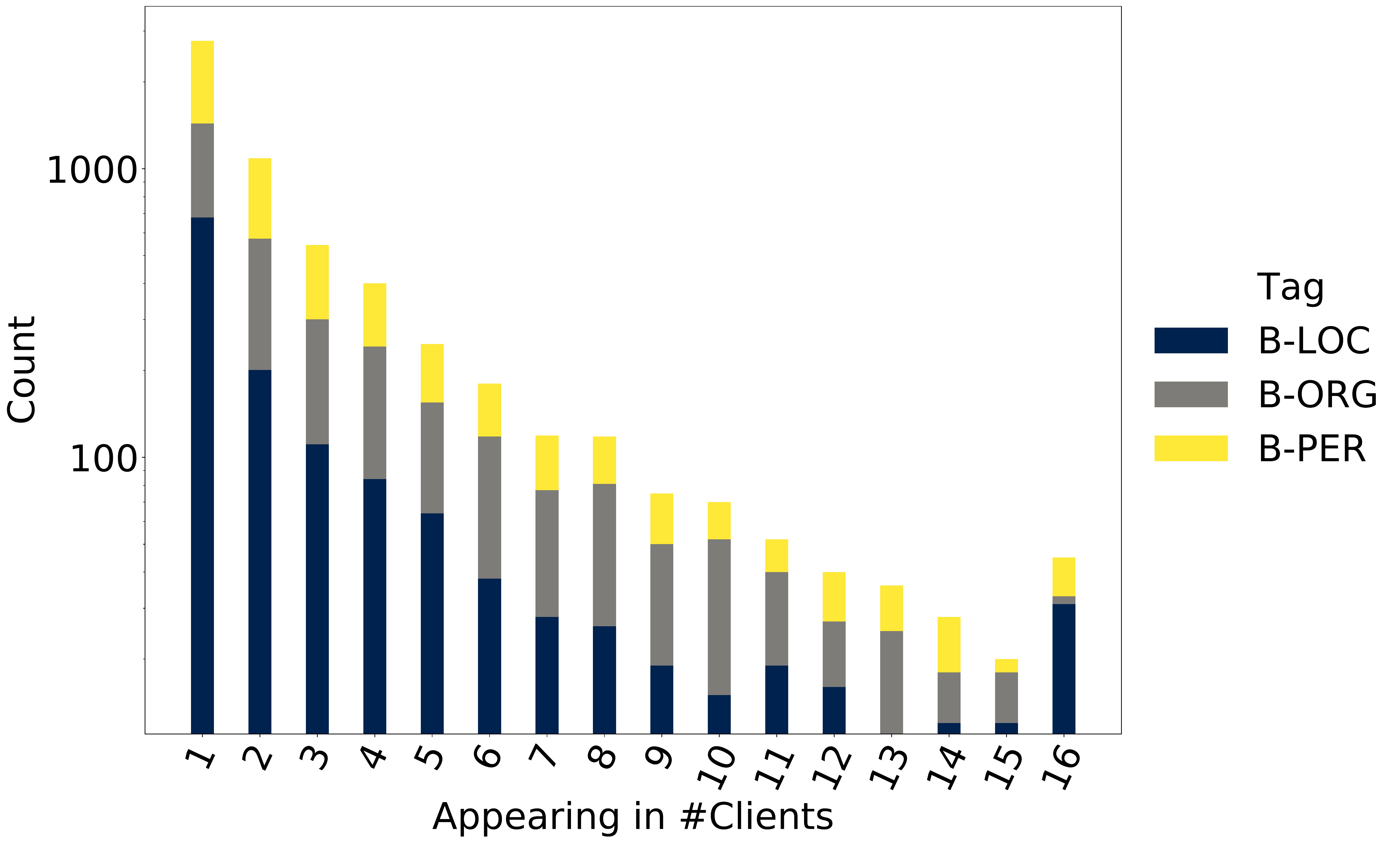}
    \label{subfig:}
  }
  
  \subfloat[32-Clients\\Uniform]{
    \centering\includegraphics[width=0.5\linewidth]{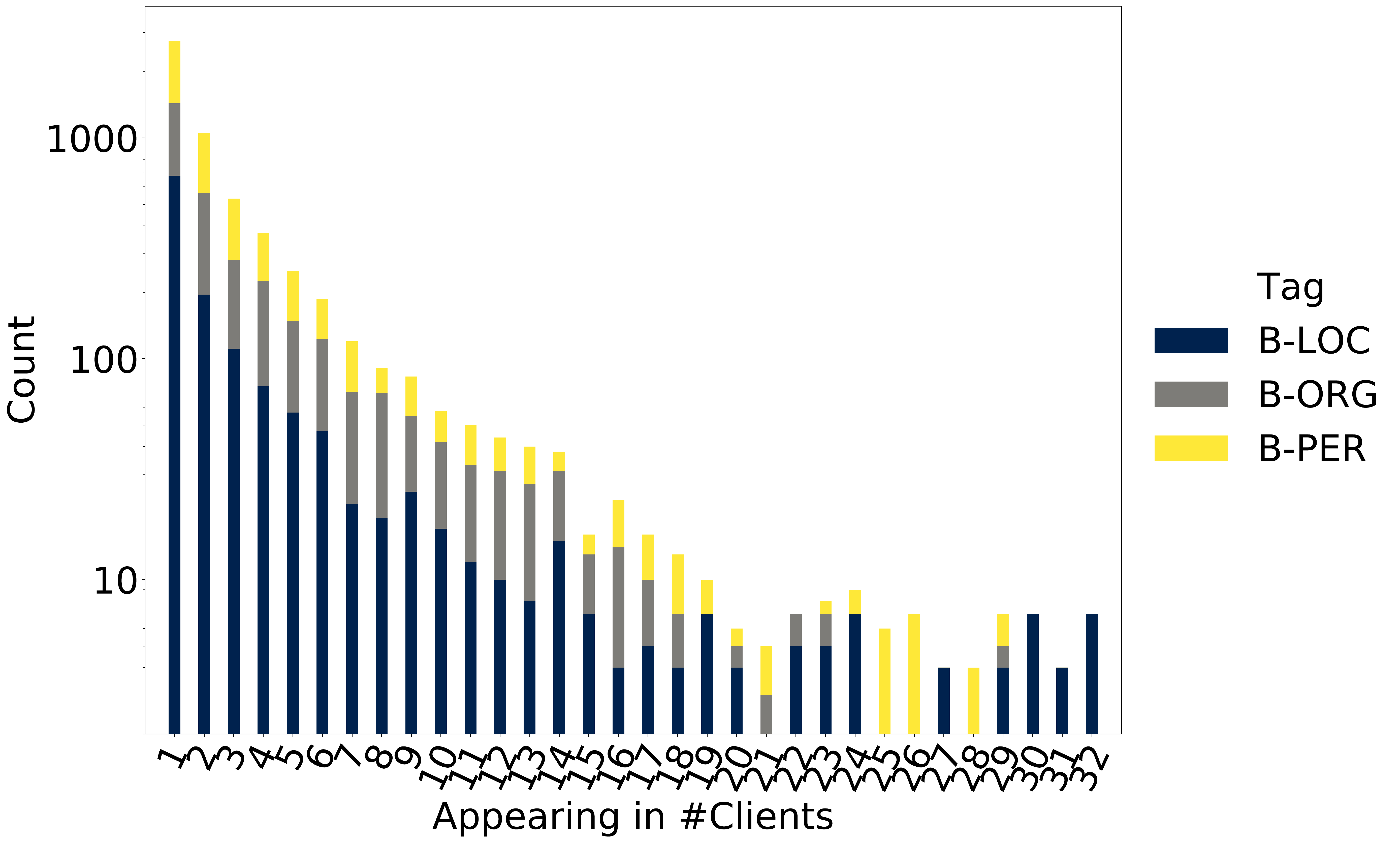}
    \label{subfig:}
  }
  \subfloat[32-Clients\\Skewed]{
    \centering\includegraphics[width=0.5\linewidth]{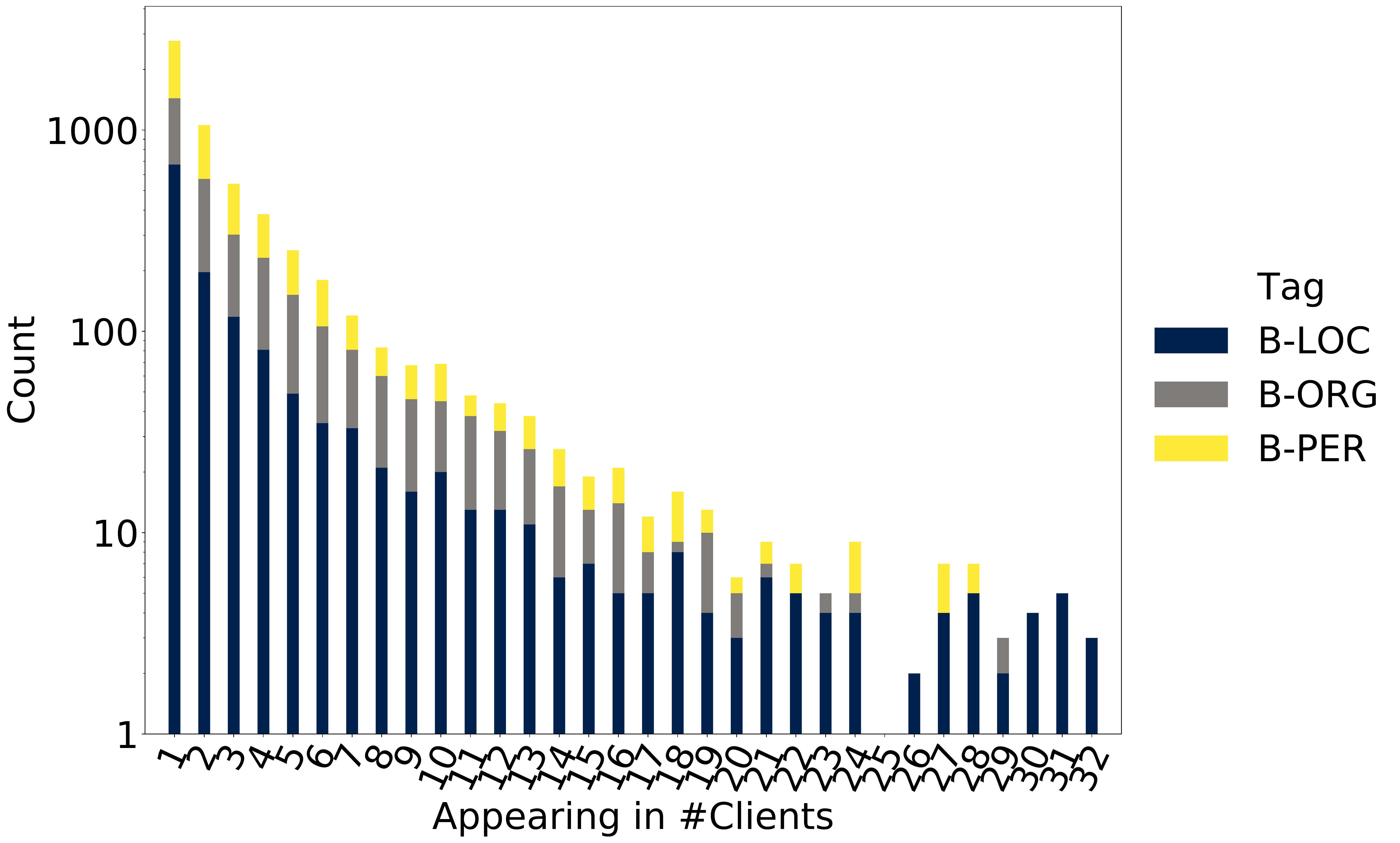}
    \label{subfig:}
  }  
  \captionsetup{}
  \caption{CoNLL-2003 dataset B-LOC, B-ORG and B-PER entity (tag) common occurrences across clients for each federation environment (log scale).}
  \label{fig:CoNLL_2003_occurences}
\end{figure}

Figure~\ref{fig:CoNLL_2003_tags_distribution} presents the total number of location, organization and person tags at each client within each federation environment.%
\footnote{The Miscellaneous (MISC) tag has a similar distribution.}
For Uniform environments, a similar amount of tags has been assigned to each client. For the Skewed environment, the distribution of tags follows a left-skewed assignment. A similar distribution pattern can be observed for tags present in only one client, which we call unique tags. Figure~\ref{fig:CoNLL_2003_unique_tags_distribution} shows the numbers of unique location, organization and person entities each client contains. 
%
Finally, Figure~\ref{fig:CoNLL_2003_occurences} shows how many clients contain each entity, that is,  how many tags appear simultaneously at K number of clients (e.g., K=5 gives how many tags appear in exactly 5 clients).


\begin{table}[htpb]
    \centering
    \begin{tabular}{cccc}
        \toprule
        {Environment} & Precision & Recall & F1-score \\
        \midrule
        Centralized & 90.91  & 90.12 & 90.51\\
        8-clients-Uni & 88.85 & 88.30 & 88.57 \\
        16-clients-Uni & 88.99 & 87.69 & 88.34 \\
        32-clients-Uni & 88.52 & 86.15 & 87.32 \\
        32-clients-Ske & 88.43 & 86.37 & 87.39 \\
        \bottomrule
    \end{tabular}
    \caption{Final learning performance of centralized vs federated models. The Clients-Uni and Clients-Ske environments refer to the federation environments with uniform and skewed assignment of training samples across clients.}
    \label{tbl:results}
\end{table}

\begin{figure}[htbp]
  \centering
  \includegraphics[width=\columnwidth]{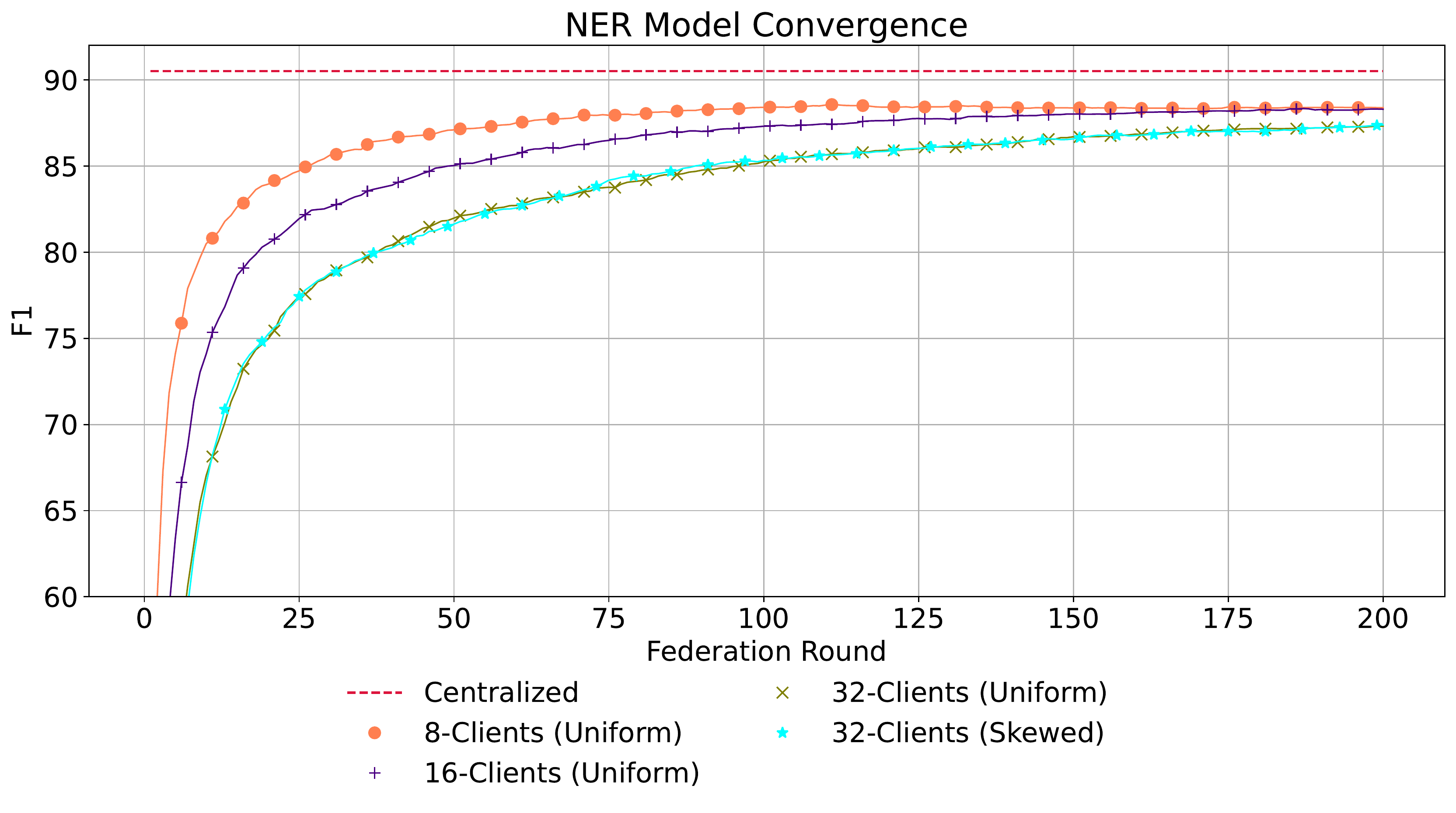}
   \caption{NER model learning performance in centralized and federated settings.}
   \label{fig:convergence}
\end{figure}

\textbf{Results.}
Table~\ref{tbl:results} shows the final learning performance of each model for each learning environment. As expected, the centralized model outperforms the federated models. However, the performance degradation is small ($\sim$2-3 percentage points). The degradation increases as the federation environments become more challenging, that is, with the number of clients and the heterogeneity of the data. As the number of clients increases, the amount of data available for local training decreases, which makes federated training harder. Nonetheless, federated learning remains effective even with larger number of clients and skewed distributions. 

Figure~\ref{fig:convergence} shows the convergence rate of the federated models, as F1-score over  federation rounds. For the centralized model we plot the best F1 score as a horizontal line. The harder the federation environment, the more training (federation) rounds the federated model needs to reach an acceptable performance.


\section{Discussion}

We have empirically shown that Federated Learning can be successfully applied in the context of NLP to train deep learning models for Named-Entity Recognition. Across the federated environments that we investigate, the final NER model can reach an acceptable performance when compared to its centralized counterpart. As the federated learning environments become more challenging due to increased statistical heterogeneity and reduced data amount per client, a moderate performance decrease occurs, and more training rounds are required for the federation to converge.

Our immediate future work focuses on training NLP models when the text at each site is private. 
A critical pre-processing step for training deep NLP models is the creation of a common vocabulary/dictionary that maps each token to a unique identifier. In a centralized setting where all the data are located at a central repository the creation of a such a vocabulary is straightforward. However, in a federated setting where the global dataset is split across multiple clients with each client having its own private local dataset, the creation of such a vocabulary is more challenging.
Due to privacy concerns, clients need to always keep their data at their original location. They cannot reveal either sentences or individual words/tokens to others. For example, consider an intelligence domain where person names and locations under investigation at a site must be protected. 
Therefore constructing a common representation that includes all federation tokens needs to be performed in a secure and private way. In our experiments, we assumed that a publicly available vocabulary exists. We are currently investigating hashing methods that can provide a global, but private, vocabulary for federated training in NLP applications.

\newpage
\bibliography{main}

\begin{thebibliography}{34}
\expandafter\ifx\csname natexlab\endcsname\relax\def\natexlab#1{#1}\fi

\bibitem[{Bellavista et~al.(2021)Bellavista, Foschini, and
  Mora}]{bellavista2021decentralised}
Paolo Bellavista, Luca Foschini, and Alessio Mora. 2021.
\newblock Decentralised learning in federated deployment environments: A
  system-level survey.
\newblock \emph{ACM Computing Surveys (CSUR)}, 54(1):1--38.

\bibitem[{Devlin et~al.(2019)Devlin, Chang, Lee, and
  Toutanova}]{devlin-etal-2019-bert}
Jacob Devlin, Ming-Wei Chang, Kenton Lee, and Kristina Toutanova. 2019.
\newblock \href {https://doi.org/10.18653/v1/N19-1423} {{BERT}: Pre-training of
  deep bidirectional transformers for language understanding}.
\newblock In \emph{Proceedings of the 2019 Conference of the North {A}merican
  Chapter of the Association for Computational Linguistics: Human Language
  Technologies, Volume 1 (Long and Short Papers)}, pages 4171--4186,
  Minneapolis, Minnesota. Association for Computational Linguistics.

\bibitem[{Ge et~al.(2020)Ge, Wu, Wu, Qi, Huang, and Xie}]{ge2020fedner}
Suyu Ge, Fangzhao Wu, Chuhan Wu, Tao Qi, Yongfeng Huang, and Xing Xie. 2020.
\newblock Fedner: Privacy-preserving medical named entity recognition with
  federated learning.
\newblock \emph{arXiv preprint arXiv:2003.09288}.

\bibitem[{Hard et~al.(2018)Hard, Rao, Mathews, Ramaswamy, Beaufays, Augenstein,
  Eichner, Kiddon, and Ramage}]{hard2018federated}
Andrew Hard, Kanishka Rao, Rajiv Mathews, Swaroop Ramaswamy, Fran{\c{c}}oise
  Beaufays, Sean Augenstein, Hubert Eichner, Chlo{\'e} Kiddon, and Daniel
  Ramage. 2018.
\newblock Federated learning for mobile keyboard prediction.
\newblock \emph{arXiv preprint arXiv:1811.03604}.

\bibitem[{Hathurusinghe et~al.(2021)Hathurusinghe, Nejadgholi, and
  Bolic}]{hathurusinghe2021privacy}
Rajitha Hathurusinghe, Isar Nejadgholi, and Miodrag Bolic. 2021.
\newblock A privacy-preserving approach to extraction of personal information
  through automatic annotation and federated learning.
\newblock \emph{arXiv preprint arXiv:2105.09198}.

\bibitem[{Hochreiter and Schmidhuber(1997)}]{Hochreiter1997LongSM}
Sepp Hochreiter and J{\"u}rgen Schmidhuber. 1997.
\newblock Long short-term memory.
\newblock \emph{Neural Computation}, 9:1735--1780.

\bibitem[{Kairouz et~al.(2021)Kairouz, McMahan, Avent, Bellet, Bennis, Bhagoji,
  Bonawitz, Charles, Cormode, Cummings et~al.}]{kairouz2021advances}
Peter Kairouz, H~Brendan McMahan, Brendan Avent, Aur{\'e}lien Bellet, Mehdi
  Bennis, Arjun~Nitin Bhagoji, Kallista Bonawitz, Zachary Charles, Graham
  Cormode, Rachel Cummings, et~al. 2021.
\newblock Advances and open problems in federated learning.
\newblock \emph{Foundations and Trends{\textregistered} in Machine Learning},
  14(1--2):1--210.

\bibitem[{Karimireddy et~al.(2020)Karimireddy, Kale, Mohri, Reddi, Stich, and
  Suresh}]{karimireddy2020scaffold}
Sai~Praneeth Karimireddy, Satyen Kale, Mehryar Mohri, Sashank Reddi, Sebastian
  Stich, and Ananda~Theertha Suresh. 2020.
\newblock Scaffold: Stochastic controlled averaging for federated learning.
\newblock In \emph{International Conference on Machine Learning}, pages
  5132--5143. PMLR.

\bibitem[{Lafferty et~al.(2001)Lafferty, McCallum, and
  Pereira}]{Lafferty2001ConditionalRF}
John~D. Lafferty, Andrew McCallum, and Fernando Pereira. 2001.
\newblock Conditional random fields: Probabilistic models for segmenting and
  labeling sequence data.
\newblock In \emph{ICML}.

\bibitem[{Lample et~al.(2016)Lample, Ballesteros, Subramanian, Kawakami, and
  Dyer}]{lample-etal-2016-neural}
Guillaume Lample, Miguel Ballesteros, Sandeep Subramanian, Kazuya Kawakami, and
  Chris Dyer. 2016.
\newblock \href {https://doi.org/10.18653/v1/N16-1030} {Neural architectures
  for named entity recognition}.
\newblock In \emph{Proceedings of the 2016 Conference of the North {A}merican
  Chapter of the Association for Computational Linguistics: Human Language
  Technologies}, pages 260--270, San Diego, California. Association for
  Computational Linguistics.

\bibitem[{Lample and Conneau(2019)}]{Lample2019CrosslingualLM}
Guillaume Lample and Alexis Conneau. 2019.
\newblock Cross-lingual language model pretraining.
\newblock In \emph{NeurIPS}.

\bibitem[{Leroy et~al.(2019)Leroy, Coucke, Lavril, Gisselbrecht, and
  Dureau}]{leroy2019federated}
David Leroy, Alice Coucke, Thibaut Lavril, Thibault Gisselbrecht, and Joseph
  Dureau. 2019.
\newblock Federated learning for keyword spotting.
\newblock In \emph{ICASSP 2019-2019 IEEE International Conference on Acoustics,
  Speech and Signal Processing (ICASSP)}, pages 6341--6345. IEEE.

\bibitem[{Li et~al.(2020)Li, Sahu, Talwalkar, and Smith}]{li2020federated}
Tian Li, Anit~Kumar Sahu, Ameet Talwalkar, and Virginia Smith. 2020.
\newblock Federated learning: Challenges, methods, and future directions.
\newblock \emph{IEEE Signal Processing Magazine}, 37(3):50--60.

\bibitem[{Li et~al.(2019)Li, Huang, Yang, Wang, and Zhang}]{li2019convergence}
Xiang Li, Kaixuan Huang, Wenhao Yang, Shusen Wang, and Zhihua Zhang. 2019.
\newblock On the convergence of fedavg on non-iid data.
\newblock In \emph{International Conference on Learning Representations}.

\bibitem[{Lin et~al.(2021)Lin, He, Zeng, Wang, Huang, Soltanolkotabi, Ren, and
  Avestimehr}]{lin2021fednlp}
Bill~Yuchen Lin, Chaoyang He, Zihang Zeng, Hulin Wang, Yufen Huang, Mahdi
  Soltanolkotabi, Xiang Ren, and Salman Avestimehr. 2021.
\newblock Fednlp: Benchmarking federated learning methods for natural language
  processing tasks.
\newblock \emph{arXiv preprint arXiv:2104.08815}.

\bibitem[{Liu et~al.(2019)Liu, Dligach, and Miller}]{liu2019two}
Dianbo Liu, Dmitriy Dligach, and Timothy Miller. 2019.
\newblock Two-stage federated phenotyping and patient representation learning.
\newblock In \emph{Proceedings of the conference. Association for Computational
  Linguistics. Meeting}, volume 2019, page 283. NIH Public Access.

\bibitem[{Liu et~al.(2021)Liu, Ho, Wang, Gao, Jin, and
  Zhang}]{liu2021federated}
Ming Liu, Stella Ho, Mengqi Wang, Longxiang Gao, Yuan Jin, and He~Zhang. 2021.
\newblock Federated learning meets natural language processing: A survey.
\newblock \emph{arXiv preprint arXiv:2107.12603}.

\bibitem[{Liu et~al.(2020{\natexlab{a}})Liu, James, Kang, Niyato, and
  Zhang}]{liu2020privacy}
Yi~Liu, JQ~James, Jiawen Kang, Dusit Niyato, and Shuyu Zhang.
  2020{\natexlab{a}}.
\newblock Privacy-preserving traffic flow prediction: A federated learning
  approach.
\newblock \emph{IEEE Internet of Things Journal}, 7(8):7751--7763.

\bibitem[{Liu et~al.(2020{\natexlab{b}})Liu, Ai, Sun, Zhang, Liu, and
  Yu}]{liu2020fedcoin}
Yuan Liu, Zhengpeng Ai, Shuai Sun, Shuangfeng Zhang, Zelei Liu, and Han Yu.
  2020{\natexlab{b}}.
\newblock Fedcoin: A peer-to-peer payment system for federated learning.
\newblock In \emph{Federated Learning}, pages 125--138. Springer.

\bibitem[{Marrero et~al.(2013)Marrero, Urbano, S{\'a}nchez-Cuadrado, Morato,
  and G{\'o}mez-Berb{\'\i}s}]{marrero2013named}
M{\'o}nica Marrero, Juli{\'a}n Urbano, Sonia S{\'a}nchez-Cuadrado, Jorge
  Morato, and Juan~Miguel G{\'o}mez-Berb{\'\i}s. 2013.
\newblock Named entity recognition: fallacies, challenges and opportunities.
\newblock \emph{Computer Standards \& Interfaces}, 35(5):482--489.

\bibitem[{Mathew et~al.(2019)Mathew, Fakhraei, and
  Ambite}]{mathew2019biomedical}
Joel Mathew, Shobeir Fakhraei, and Jos{\'e}~Luis Ambite. 2019.
\newblock Biomedical named entity recognition via reference-set augmented
  bootstrapping.
\newblock \emph{arXiv preprint arXiv:1906.00282}.

\bibitem[{McMahan et~al.(2017)McMahan, Moore, Ramage, Hampson, and
  y~Arcas}]{mcmahan2017communication}
Brendan McMahan, Eider Moore, Daniel Ramage, Seth Hampson, and Blaise~Aguera
  y~Arcas. 2017.
\newblock Communication-efficient learning of deep networks from decentralized
  data.
\newblock In \emph{Artificial intelligence and statistics}, pages 1273--1282.
  PMLR.

\bibitem[{Mitra et~al.(2021)Mitra, Jaafar, Pappas, and
  Hassani}]{mitra2021linear}
Aritra Mitra, Rayana Jaafar, George Pappas, and Hamed Hassani. 2021.
\newblock Linear convergence in federated learning: Tackling client
  heterogeneity and sparse gradients.
\newblock \emph{Advances in Neural Information Processing Systems}, 34.

\bibitem[{Pennington et~al.(2014)Pennington, Socher, and
  Manning}]{pennington-etal-2014-glove}
Jeffrey Pennington, Richard Socher, and Christopher Manning. 2014.
\newblock \href {https://doi.org/10.3115/v1/D14-1162} {{G}lo{V}e: Global
  vectors for word representation}.
\newblock In \emph{Proceedings of the 2014 Conference on Empirical Methods in
  Natural Language Processing ({EMNLP})}, pages 1532--1543, Doha, Qatar.
  Association for Computational Linguistics.

\bibitem[{Reddi et~al.(2020)Reddi, Charles, Zaheer, Garrett, Rush,
  Kone{\v{c}}n{\`y}, Kumar, and McMahan}]{reddi2020adaptive}
Sashank~J Reddi, Zachary Charles, Manzil Zaheer, Zachary Garrett, Keith Rush,
  Jakub Kone{\v{c}}n{\`y}, Sanjiv Kumar, and Hugh~Brendan McMahan. 2020.
\newblock Adaptive federated optimization.
\newblock In \emph{International Conference on Learning Representations}.

\bibitem[{Rieke et~al.(2020)Rieke, Hancox, Li, Milletari, Roth, Albarqouni,
  Bakas, Galtier, Landman, Maier-Hein et~al.}]{rieke2020future}
Nicola Rieke, Jonny Hancox, Wenqi Li, Fausto Milletari, Holger~R Roth, Shadi
  Albarqouni, Spyridon Bakas, Mathieu~N Galtier, Bennett~A Landman, Klaus
  Maier-Hein, et~al. 2020.
\newblock The future of digital health with federated learning.
\newblock \emph{NPJ digital medicine}, 3(1):1--7.

\bibitem[{Sang and De~Meulder(2003)}]{sang2003introduction}
Erik~F Sang and Fien De~Meulder. 2003.
\newblock Introduction to the {CoNLL-2003} shared task: Language-independent
  named entity recognition.
\newblock \emph{arXiv preprint cs/0306050}.

\bibitem[{Sheller et~al.(2020)Sheller, Edwards, Reina, Martin, Pati, Kotrotsou,
  Milchenko, Xu, Marcus, Colen et~al.}]{sheller2020federated}
Micah~J Sheller, Brandon Edwards, G~Anthony Reina, Jason Martin, Sarthak Pati,
  Aikaterini Kotrotsou, Mikhail Milchenko, Weilin Xu, Daniel Marcus, Rivka~R
  Colen, et~al. 2020.
\newblock Federated learning in medicine: facilitating multi-institutional
  collaborations without sharing patient data.
\newblock \emph{Scientific reports}, 10(1):1--12.

\bibitem[{Stremmel and Singh(2021)}]{stremmel2021pretraining}
Joel Stremmel and Arjun Singh. 2021.
\newblock Pretraining federated text models for next word prediction.
\newblock In \emph{Future of Information and Communication Conference}, pages
  477--488. Springer.

\bibitem[{Stripelis and Ambite(2021)}]{stripelis2021semi}
Dimitris Stripelis and Jos{\'e}~Luis Ambite. 2021.
\newblock Semi-synchronous federated learning.
\newblock \emph{arXiv preprint arXiv:2102.02849}.

\bibitem[{Wang et~al.(2021)Wang, Charles, Xu, Joshi, McMahan, Al-Shedivat,
  Andrew, Avestimehr, Daly, Data et~al.}]{wang2021field}
Jianyu Wang, Zachary Charles, Zheng Xu, Gauri Joshi, H~Brendan McMahan, Maruan
  Al-Shedivat, Galen Andrew, Salman Avestimehr, Katharine Daly, Deepesh Data,
  et~al. 2021.
\newblock A field guide to federated optimization.
\newblock \emph{arXiv preprint arXiv:2107.06917}.

\bibitem[{Wang et~al.(2020)Wang, Liu, Liang, Joshi, and
  Poor}]{wang2020tackling}
Jianyu Wang, Qinghua Liu, Hao Liang, Gauri Joshi, and H~Vincent Poor. 2020.
\newblock Tackling the objective inconsistency problem in heterogeneous
  federated optimization.
\newblock \emph{Advances in neural information processing systems},
  33:7611--7623.

\bibitem[{Yang et~al.(2019)Yang, Liu, Chen, and Tong}]{yang2019federated}
Qiang Yang, Yang Liu, Tianjian Chen, and Yongxin Tong. 2019.
\newblock Federated machine learning: Concept and applications.
\newblock \emph{ACM Transactions on Intelligent Systems and Technology (TIST)},
  10(2):1--19.

\bibitem[{Yang et~al.(2018)Yang, Andrew, Eichner, Sun, Li, Kong, Ramage, and
  Beaufays}]{yang2018applied}
Timothy Yang, Galen Andrew, Hubert Eichner, Haicheng Sun, Wei Li, Nicholas
  Kong, Daniel Ramage, and Fran{\c{c}}oise Beaufays. 2018.
\newblock Applied federated learning: Improving google keyboard query
  suggestions.
\newblock \emph{arXiv preprint arXiv:1812.02903}.

\end{thebibliography}




\end{document}